\documentclass[conference]{IEEEtran}
\usepackage{subcaption}
\IEEEoverridecommandlockouts
\usepackage{cite}
\usepackage{amsmath,amssymb,amsfonts}
\usepackage{algorithmic}
\usepackage{graphicx}
\usepackage{parskip}
\usepackage{textcomp}
\usepackage{xcolor}
\usepackage{subcaption}

\usepackage{booktabs}
\usepackage{bigstrut}
\usepackage{multirow}
\usepackage{multicol}
\usepackage[T1]{fontenc}

\def\BibTeX{{\rm B\kern-.05em{\sc i\kern-.025em b}\kern-.08em
    T\kern-.1667em\lower.7ex\hbox{E}\kern-.125emX}}

\newcommand{\etal}{\textit{et al}.}
\newcommand{\ie}{\textit{i}.\textit{e}.}
\newcommand{\eg}{\textit{e}.\textit{g}.}

\newcommand{\authorskip}{\hspace{2.8mm}}
\newcommand{\emailskip}{\hspace{4mm}}

\begin{document}

\title{sDREAMER: Self-distilled Mixture-of-Modality-Experts Transformer for Automatic Sleep Staging}

\author{
 Jingyuan Chen$^{1}$ \authorskip Yuan Yao$^{1}$ \authorskip Mie Anderson$^{2}$ \authorskip
 Natalie Hauglund$^{2}$ \authorskip Celia Kjaerby$^{2}$  \\ \authorskip Verena Untiet$^{2}$ \authorskip Maiken Nedergaard$^{2}$  \authorskip Jiebo Luo$^{1}$ \\[2mm]
 $^{1}$University of Rochester  \qquad $^{2}$University of Copenhagen \\
 \texttt{\small jchen157@u.rochester.edu} \emailskip
 \texttt{\small yyao39@ur.rochester.edu} \emailskip
 \texttt{\small jluo@cs.rochester.edu}\\[0mm]
 \texttt{\small \{mieandersen,natalie.hauglund,celia.kjaerby,verena,nedergaard\}@sund.ku.edu}
}

\maketitle

\begin{abstract}
Automatic sleep staging based on electroencephalography (EEG) and electromyography (EMG) signals is an important aspect of sleep-related research. 
Current sleep staging methods suffer from two major drawbacks. 
First, there are limited information interactions between modalities in the existing methods.
Second, current methods do not develop unified models that can handle different sources of input. 
To address these issues, we propose a novel sleep stage scoring model sDREAMER, which emphasizes cross-modality interaction and per-channel performance. Specifically, we develop a mixture-of-modality-expert (MoME) model with three pathways for EEG, EMG, and mixed signals with partially shared weights. We further propose a self-distillation training scheme for further information interaction across modalities. Our model is trained with multi-channel inputs and can make classifications on either single-channel or multi-channel inputs. Experiments demonstrate that our model outperforms the existing transformer-based sleep scoring methods for multi-channel inference. For single-channel inference, our model also outperforms the transformer-based models trained with single-channel signals.
\end{abstract}


\section{Introduction}
\label{sec:intro}

Sleep is one of the most basic animal behaviours with wide biological implications.
Classifying sleep into stages (typically Wakefulness, REM and non-REM) is an important aspect of sleep-related research~\cite{stephansen2018neural}.
Currently, one of the most common automated sleep staging techniques is based on electrophysiological time-series data.
Polysomnography (PSG) by experts requires specialized knowledge of sleep architecture, as well as considerable time (\ie experts have to observe the whole sleep process, which lasts about several hours), thus validating the necessity of automatic sleep staging. 

In recent years, deep-learning-based methods have shown potential in sleep staging.
Most of these works involve the electroencephalography (EEG) time series signals signals~\cite{biswal2017sleepnet, phan2019seqsleepnet, huang2022improved}.
However, it has been shown that sleep staging with only EEG signals is insufficient for sleep classification (typically in distinguishing REM sleep and non-REM sleep)~\cite{kim2018automatic, andreotti2018multichannel}, thus necessitating multi-modality sleep staging (\ie\ staging with multiple types of input signals)~\cite{phan2021xsleepnet, jia2021salientsleepnet, akada2021deep, pradeepkumar2022towards}.
Although these multi-modality methods have improved the performance of sleep staging, there are two major drawbacks.

\begin{figure}
    \centering
    \includegraphics[width=0.85\linewidth]{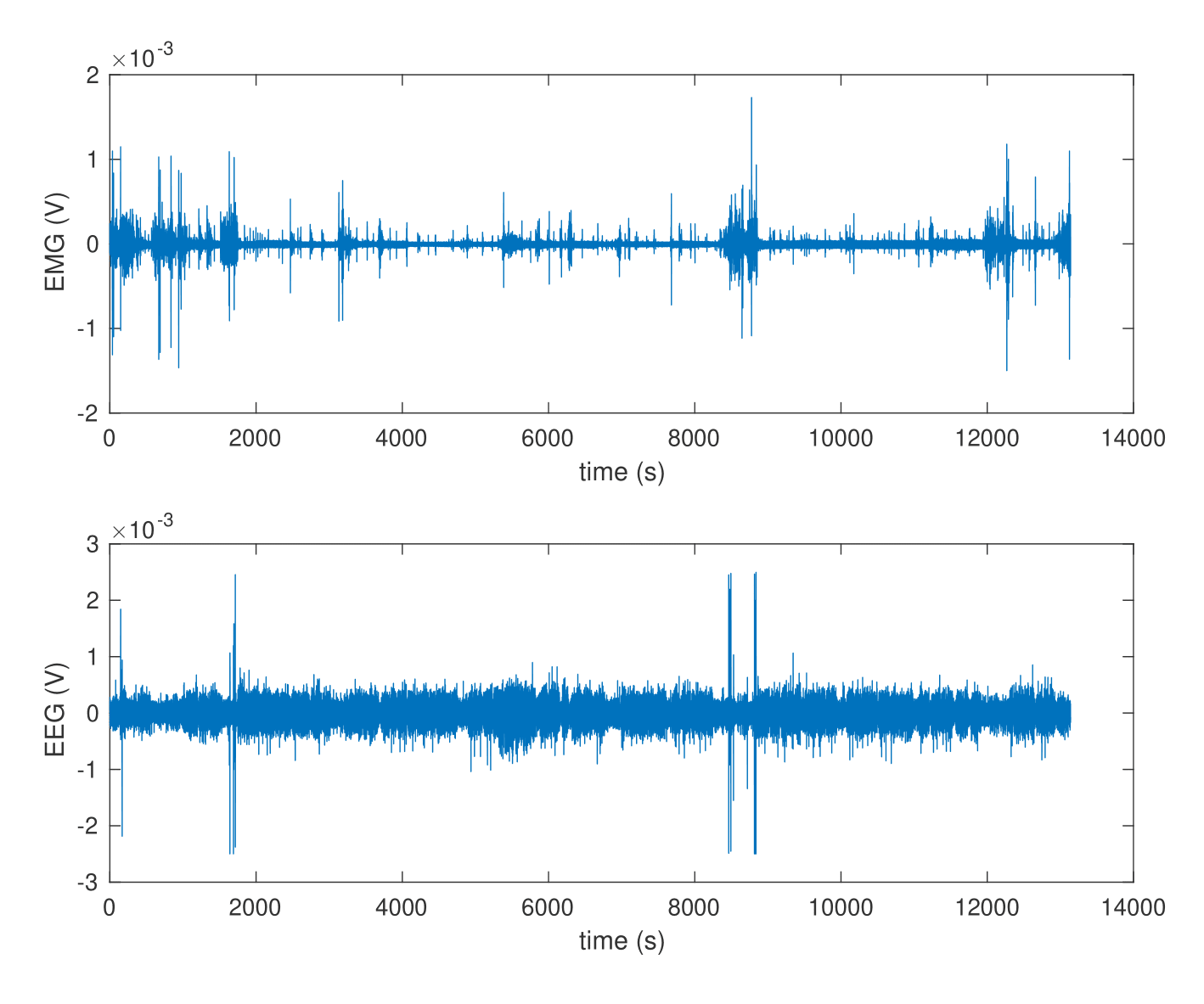}
    \caption{The raw EMG and EEG time-series signals.}
    \label{fig:raw}
\end{figure}

First, previous methods suffer from limited information interactions between modalities.
While previous studies have attempted to develop multi-modal models that integrate information across different modalities, they ignore the possible interrelationship of different electrophysiological signals. Specifically, these models embed different signal inputs into separate feature spaces and use a late fusion scheme to generate classification results. 
Since the electrophysiological signals are quite noisy (see Fig.~\ref{fig:raw}), the cross-reference from multiple input signals can be essential in improving the robustness of automatic sleep staging models.

Second, previous methods have not developed unified models that can handle different sources of input, such as multi-channel and single-channel signals.
For instance, a model trained on EEG and EMG signals cannot perform reasonable inferences on new data with only EEG signals.
This limitation is brought naturally since such models focus on multi-channel joint embedding, but neglect the quality of per-channel embedding. As a result, the performance of these models is restricted, and their potential applications are hindered. Therefore, there is a need for a unified model that can handle different sources of input and improve the performance of automatic sleep staging models.

To address these problems, this study proposes a novel sleep stage scoring model that emphasizes the cross-modality alignment and per-channel performance.
To achieve this, our model introduces the mixture-of-modality-experts (MoME) transformer.
More specifically, we develop a model consisting of three pathways for EEG, EMG, and mixed signals with partially shared weights. Specifically, the attention weights are shared across the three paths, and the weights of the feed-forward networks are not fully shared (see Section~\ref{sec:method} for details).
The partially shared weights implicitly instruct our model to align across different modalities.
The three pathways ensure that our model can promote high-quality per-channel performance.
Furthermore, we propose a self-distillation method to ensure better information interaction across modalities.
We build an epoch MoME transformer for one-to-one sleep staging and a sequence MoME transformer for many-to-many sleep staging. 
Although our model is trained with multi-channel time-series signals (\ie\ EEG and EMG), our model can accept single-channel (\ie\ EEG or EMG) as well as multi-channel time-series signal as input in the inference phase.

We demonstrate the effectiveness of our proposed self-distilled MoME transformer, referred to as sDREAMER (\underline{s}elf-\underline{d}istilled
Mixtu\underline{re}-of-Mod\underline{a}lity-Experts Transfor\underline{mer}), through experiments on a public mice sleep dataset~\cite{kjaerby2022memory} labeled with sleep stages by our expert. For multi-channel sleep staging, our epoch MoME model and sequence MoME model outperform the transformer-based automatic sleep scoring method by significant margins, respectively. 
For single-channel sleep staging, our sequence MoME model also produces better inference results than the transformer-based models trained with single-channel signals.

In summary, the main contributions of our work can be summarized as follows:
\begin{itemize}
\item We propose the unified sleep staging framework sDREAMER that can handle either single-channel or multi-channel input based on the MoME transformer.
\item We propose a self-distillation method for our proposed MoME transformer to ensure the multi-modality information interaction across different pathways (\ie EEG, EMG, and mixed).
\item We demonstrate through extensive experiments that our proposed model is effective for both single-channel and multi-channel sleep staging.
\end{itemize}

\section{Related Work}
\subsection{Deep-learning-based Sleep Staging}
The success of deep learning has inspired several deep-learning-based sleep staging algorithms based on electrophysiological signals.
Basically, there are two ways to deal with the raw data. One way is to directly learn from the unprocessed signals~\cite{supratak2017deepsleepnet, schwabedal2018automated, PATHAK2021102038}. Another way is to transfer them to spectrograms and design models to learn from the spectrograms using computer vision-based techniques~\cite{vilamala2017deep, miladinovic2019spindle}.
Although introducing spectrograms may produce more interpretable visual understanding for humans, Phan~\etal ~have shown that introducing spectrogram inputs in addition to raw signal inputs may actually lead to a performance drop~\cite{phan2021xsleepnet}. Therefore, we focus on automatic sleep staging based on raw electrophysiological signals in our work. 

Most automatic sleep staging works involve the time-series EEG signals~\cite{biswal2017sleepnet, phan2018dnn, sors2018convolutional, phan2018automatic, phan2019seqsleepnet, eldele2021attention, huang2022improved, phan2022sleeptransformer}.
However, previous works have shown that sleep staging with only EEG signals is insufficient for sleep classification (typically in distinguishing REM sleep and non-REM sleep)~\cite{kim2018automatic, andreotti2018multichannel}, thus validating the necessity of multi-modality sleep staging (\ie staging with multiple types of input signals)~\cite{phan2021xsleepnet, jia2021salientsleepnet, akada2021deep, pradeepkumar2022towards}.
One common way of modeling the different types of input signals is the late fusion strategy~\cite{mikkelsen2018personalizing, jia2022multi}.
Despite their success in introducing different time-series signals into sleep staging, one major problem of these methods is the lack of information interaction across modalities. 
Although Jathurshan~\etal ~have proposed cross-modal attention~\cite{pradeepkumar2022towards}, the shallow design is insufficient for information interactions.

\subsection{Transformer-based Cross-modality Learning}
Transformer~\cite{vaswani2017attention} is a deep-learning framework that can handle sequence input. The recent success in vision-language has demonstrated the power of transformer in cross-modality learning~\cite{radford2021learning, li2021align, wang2022image}.
Typically, there are two types of structures for cross-modality learning: 1) fusion encoder, where different input modalities share the same encoder structure~\cite{tan2019lxmert, zhang2021vinvl} and 2) dual encoder, where different input modalities have different encoder structure but are mixed together in the feature space for cross-modal interactions~\cite{jia2021scaling, wang2021distilled, zhang2021cross}. Bao~\etal ~suggest that mixture-of-modality-experts transformer can facilitate cross-modality learning efficiently and competitively~\cite{bao2022vlmo}.
Despite the effectiveness of cross-modality learning, most works on sleep staging ignore such techniques for modeling different types of input signals.

\subsection{Self Distillation}
Self distillation is a type of knowledge distillation where the teacher network and the student network share the same model. 
In one type of self-distillation methods, knowledge can be transferred from the earlier epochs to the later epochs~\cite{yang2019snapshot, zhao2021knowledge, shen2022self}.
Knowledge can also be transferred from deeper parts of a neural network to the more shallow parts~\cite{zhang2019your, hou2019learning}.
Other self-distillation methods focus on the data and label augmentation~\cite{zhang2020self, dong2022maskclip} or weak labels~\cite{zhang2021perturbed, andonian2022robust}.
The effectiveness of self-distillation has been analyzed and validated by some works~\cite{mobahi2020self, allen2020towards}.
Recently, Wang~\etal ~have proposed to distill across vision and language domains~\cite{wang2021distilled}. Inspired by their work, we propose a novel knowledge distillation scheme in the MoME module to distill across signal modalities for sleep staging.
\section{Data Collection and Preprocessing}

\subsection{Data Collection}
This section presents the data collection procedures of the 
mouse sleep staging dataset. The dataset is a publicly available mice sleep dataset~\cite{kjaerby2022memory}, while our experts manually labeled the dataset with sleep stages. All the sleep monitoring data are recorded from wildtype C57BL/6 mice subjects including both male and female ones. In particular, the EEG and EMG signals are collected when the subjects are placed in the chamber room. To denoise the collected EEG signals, experts apply filters including a high-pass filter at 1 Hz and a low-pass filter at 100 Hz. For the EMG signals, they apply a high-pass filter at 10 Hz and a low-pass filter at 100 Hz. In addition, a notch filter of 50 Hz is applied to eliminate the power line noise. Based on these raw data and the videos of sleeping mice, our experts annotate each collected signal epoch with its corresponding sleep stage. Specifically, wake stages show high muscle tonus and a high-frequency, low-amplitude EEG pattern. SWS sleep shows no muscle tonus and low-frequency, high-amplitude EEG pattern. REM sleep also shows no muscle tonus but with high frequency, low-amplitude EEG. A detailed description of the dataset can be found in Section~\ref{sec:dataset}.

\subsection{Data Preprocessing}
The dataset mainly contains the EEG and EMG signal records collected from sleeping mice, as well as the sleep stage labels for every second. These raw data may suffer from ambiguity in labels, data noise and excessive sequence length. Data preprocessing is applied to the raw data, including irregularity removal, subject-wise normalization, and temporal slicing.

\textbf{Irregularity Removal.} During the whole span of mice sleep, there exists some time periods when experts are not sure which stage these periods belong to. Such periods are not labeled, which causes data irregularity. To handle these missing values, a neurally controlled differential equation approach~\cite{kidger2020neural} was considered. However, since missing values only occurs in a small percentage of the total data, it is unnecessary to apply this method to the large electrophysiological signal dataset. Instead, we remove missing values from the time series during loss calculation.

\textbf{Subject-wise Normalization.} Raw signal normalization is performed using a subject-wise normalization approach, which normalizes the signals of the same subject with their mean and standard deviation. This method facilitates model learning as electrophysiological signal features often differ greatly in amplitude among individual subjects.

\textbf{Temporal Slicing.} To obtain input data of a fixed time span, temporal slicing is performed. As our model aims to offer accurate prediction at the second level, the input signals are sliced into per-second signals. The sampling frequency for both signals are equal, denoted as $T$. Since we choose one second as the slicing time span, the sliced EEG or EMG time-series signal can be denoted as $x^{\text{EEG}} \in \mathbb{R}^{T\times 1}$ or $x^{\text{EMG}} \in \mathbb{R}^{T\times 1}$, respectively.
\section{Methodology}
\label{sec:method}
In this study, we propose sDREAMER, an efficient multi-modal learning framework to stage sleep using a self-distilled mixture-of-modality-experts transformer. We first establish the problem hierarchically into epoch and sequence settings. Then we present the data input under these two settings. Afterward, the MoME module is introduced as the core for the unified multi-modal learning framework. The MoME module also serves as the foundation for our proposed Epoch sDREAMER and Sequence sDREAMER. Both models support multi-modal and mono-modal input cases. To further enhance the performance of mono-modality input, we propose a self-distillation pipeline that employs a multi-modal expert to guide the mono-modality expert.

\begin{figure*}[t!]
    \centering
    \includegraphics[width=0.95\linewidth]{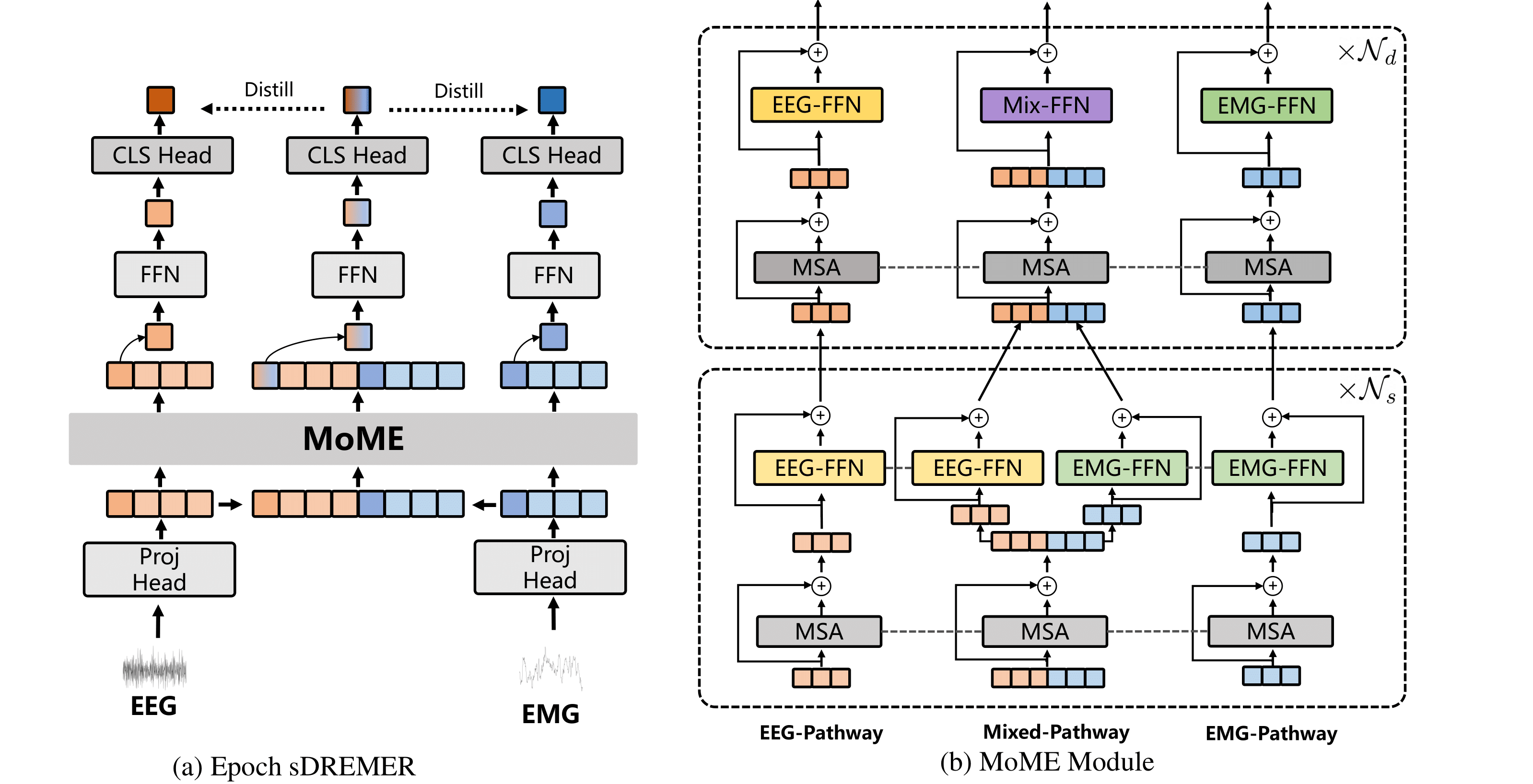}
    \caption{The overall structure of the mixture-of-modality-experts module and the epoch sDREAMER model.}
    \label{fig:mome}
\end{figure*}

\subsection{Problem Setup}
Sleep stage scoring in mice is commonly approached as a multi-class classification task, requiring the model to make predictions according to the electrophysiological signals. Here we introduce two different settings for sleep stage scoring: epoch and sequence setup.

\textbf{Epoch Setting.}
For sleep staging, the smallest annotation window (one second for our dataset) of a signal is referred to as an epoch. As previously defined, the sliced signals $x^{\text{EEG}}$ or $x^{\text{EMG}}$ belongs to $\mathbb{R}^{T\times 1}$, 
where $T$ is the size of time dimension, representing number of signal data points in an epoch.
Since the input signals can be single-channeled or multi-channeled, an additional modality dimension is introduced, and the epoch extends to 
$x \in \mathbb{R}^{M \times T\times 1}$, where $M$ is the size of modality dimension. 
The epoch setting can then be formulated as producing a sleep stage prediction
$\hat{y}$ for each epoch signal $x \in \mathbb{R}^{M \times T \times 1}$.

\textbf{Sequence Setting.}
The sequence setting differs in considering multiple epochs as input and producing multiple predictions as output. A sequence is a larger window within the whole signal trace. It consists of multiple consecutive epochs. A $K$-size sequence is denoted as $\left(x_i\right)_{i=1}^{K}$ where $x_i$ is the $i$-th epoch within the sequence. The sleep staging for a sequence is defined as the process of using a set of epoch signals, represented as $\left(x_i\right)_{i=1}^{K}$, to predict a corresponding set of predictions, denoted as $\left(\hat{y_i}\right)_{i=i}^{K}$ by the model. These two settings are also referred to as one-to-one and many-to-many predictions conventionally\cite{phan2018joint} \cite{pradeepkumar2022towards}.

\subsection{Data Input}
\textbf{Patching Operator}
 To process the epoch signal $x_{i}$ in a $K$-size sequence $\left(x_1,x_2,\cdots, x_{K}\right)$, a patch operator is used to divide each epoch signal into a series of non-overlapping windows of equal length $W$. This operation generates a patched signal $x_{i} \in \mathbb{R}^{P \times W}$, where $P = \lfloor \frac{T}{W} \rfloor$. 

The patch operator reduces the number of tokens for transformer input by a factor of $P$, resulting in a significant decrease in the computation complexity for multi-head self-attention by a factor of $P^2$, thereby enabling more efficient modeling.

\textbf{Initial Representation}
Given an EEG and EMG epoch $x$, we apply the aforementioned patching operator to obtain their signal patches. The resulting patches are then embedded in latent spaces to form intermediate initial tokens $\boldsymbol{I}_{0}^{m}$ as in Equation~\ref{eq:initial representations}, where $m$ denotes the modality, $m_{i}$ denotes the $i$-th epoch of the modality, and $\boldsymbol{E}_\text{trans}^{m}$ is the linear transformation matrix of this modality. 

\begin{equation}
\label{eq:initial representations}
\begin{aligned}
    \boldsymbol{I}_{0}^{m(i)} = \left[m_{i}^{1}\boldsymbol{E}_\text{trans}^{m}; 
    m_{i}^{2}\boldsymbol{E}_\text{trans}^{m};\cdots  m_{i}^{P}\boldsymbol{E}_\text{trans}^{m} \right], \boldsymbol{E}_\text{trans}^{m} \in \mathbb{R}^{W \times D}
\end{aligned}
\end{equation}
Following the Vision Transformer\cite{dosovitskiy2020image}, we initialize two $\left[\text{CLS}\right]$ tokens for EEG and EMG semantic representation within an epoch.
To preserve the temporal and modality relationships among patch tokens, we introduced joint attribute encoding which combines learnable positional and modality encodings. The intermediate initial representations are concatenated with the $\left[\text{CLS}\right]$ tokens, and the attribute encoding is then added elementwise to the combined representation as shown in Equation~\ref{eq:attribute encoding}.

\begin{equation}
\label{eq:attribute encoding}
\begin{aligned}
    \boldsymbol{E}_\text{attr}^{m} &= \boldsymbol{E}_\text{pos}^{m} + \boldsymbol{E}_\text{mod}^{m}, \quad \quad \boldsymbol{E}_\text{pos}^{m} \, ,
    \boldsymbol{E}_\text{mod}^{m} \in \mathbb{R}^{P \times D} \\
    \boldsymbol{T}_{0}^{m\left(i\right)} &= \text{Concat}\left(\boldsymbol{T}_\text{cls}^{m};
    \boldsymbol{I}_{0}^{m\left(i\right)}\right) + \boldsymbol{E}_\text{attr}^{m}
\end{aligned}
\end{equation}

Here, $\boldsymbol{T}_\text{cls}^{m}$, $\boldsymbol{E}_\text{pos}^{m}$, $\boldsymbol{E}_\text{mod}^{m}$ and $ \boldsymbol{T}_{0}^{m(i)}$ represents the $\left[\text{CLS}\right]$ token, positional encoding, modality encoding and final initial representations for modality $m$.
Afterward, the initial tokens for EEG and EMG at $i$-th epoch are created and denotes as $\boldsymbol{T}_{0}^\text{eeg(i)}$ and $\boldsymbol{T}_{0}^\text{emg(i)}$, respectively.

\begin{figure*}
    \centering
    \includegraphics[width=1.0\textwidth]{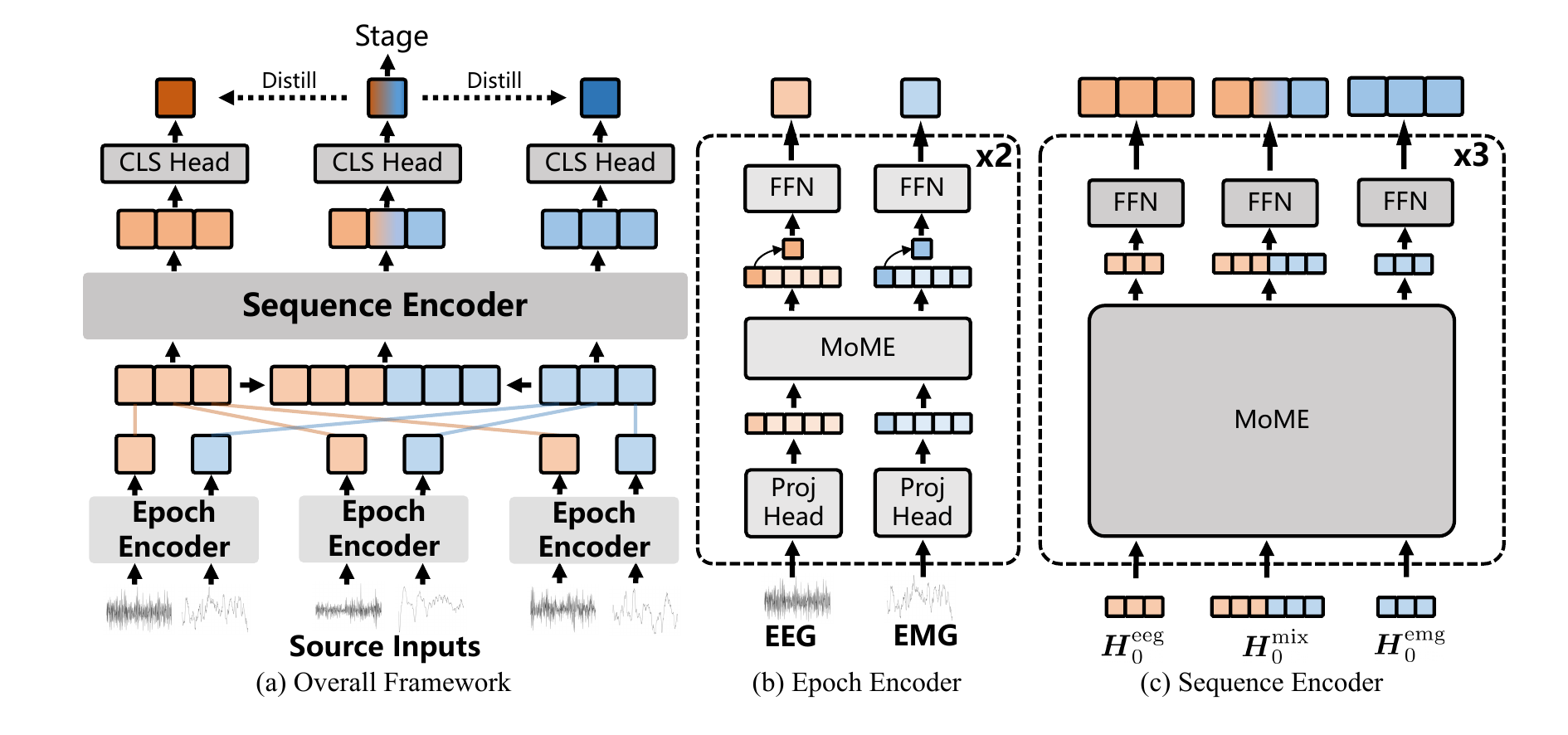}
    \caption{Overview of the sequence sDREAMER model. The structure of MoME module has already been illustrated in Fig.~\ref{fig:mome}.}
    \label{fig:Seq_MoME}
\end{figure*} 

\subsection{Mixture-of-Modality-Experts Module}
The success of mixture-of-modality-experts (MoME)~\cite{bao2022vlmo} in the visual language understanding task led us to investigate its potential strength in mice sleep scoring. In our paper, the MoME module is a module of three pathways for EEG, EMG and mix data, with partially shared weights. Our proposed epoch sDREAMER and sequence sDREAMER models are all built upon the MoME module to learn the contextual information within a given signal window. In this section, we discuss a unified token processing framework for the MoME module that applies to both models. 

Specifically, a MoME module is composed of multiple MoME layers, as shown in Fig.~\ref{fig:mome}.  Each MoME layer is an efficient variant of the transformer layer but differs in the use of modality-agnostic shared multi-head self-attention (MSA) and modality-aware feedforward neuron networks (FFN). These modality-aware FNNs that capture modality-specific latent features are conventionally referred to as modality experts. To specify, our sleep staging model has three modality experts: EEG, EMG, and mix experts. By designing different forward propagation pathways that pass through these modality experts, the MoME module is able to generate both mono-modal and multi-modal representations. 

Given the initial representations of $N$ tokens from a modality $m$ denoted as $\boldsymbol{T}_{0}^{m}$ , a pathway of $L$ layers is represented as follows:
\begin{align}
    \boldsymbol{T}_{0}^{m} &= \left[\boldsymbol{t}_{0}^{1}; \boldsymbol{t}_{0}^{2}; \cdots ; \boldsymbol{t}_{0}^{N}\right]^{m}\\
    \boldsymbol{I}_{\ell}^{m} &= \text{MSA}\left(\text{LN} \left(\boldsymbol{T}_{\ell-1}^{m}\right)\right) + \boldsymbol{T}_{\ell-1}^{m}, \quad \quad {\ell}&=1 \ldots L \\
    \boldsymbol{T}_{\ell}^{m} &= \boldsymbol{\psi}\left({m},{\ell}\right)\left(\text{LN} \left(\boldsymbol{I}_{\ell}^{m}\right)\right) + \boldsymbol{T}_{\ell}^{m^{\prime}}, \quad \quad {\ell}&=1 \ldots L 
\end{align}

Here, the $\boldsymbol{\psi}\left({m},{\ell}\right)$ is a mapping function from the layer-modality joint space $\boldsymbol{D}$ to the modality experts space $\boldsymbol{F}$ as shown in Equation \ref{eq:mome-ffn}. In other words, the activation of the modality experts is determined by the input modality and layer index. Such a mapping design allows the model to handle various inputs and produce multiple outputs with a unified model. Fig.~\ref{fig:mome} provides a graphical illustration of this mapping function. 

\begin{equation}
\label{eq:mome-ffn}
\begin{aligned}
    \boldsymbol{\psi}&\left({m},{\ell}\right): \boldsymbol{D}\to \boldsymbol{F}, \boldsymbol{D}=\left\{(m,\ell)|m \in \boldsymbol{M}, \ell \in \left\{i\right\}_{i=1}^{L} \right\}\\
    \boldsymbol{F} &= \left\{\text{FFN}^\text{eeg}, \text{FFN}^\text{emg}, \text{FFN}^\text{mix} \right\}, \boldsymbol{M} = \left\{\text{eeg}, \text{emg}, \text{mix}\right\}
\end{aligned}
\end{equation}

\textbf{EEG and EMG Pathways.}
Our MoME modules consist of two mono-modal pathways, namely the EEG-pathway and the EMG-pathway. The EEG/EMG pathway takes the input tokens corresponding to each modality and feeds them through each MoME layer to learn mono-modal contextual information. In each MoME layer, the representations first pass through the multi-head self-attention layer to learn the temporal dependencies between tokens. Next, the representations are projected into modality-specific latent space. Finally, the token representations $\boldsymbol{T}_{L}^\text{eeg}$ and $\boldsymbol{T}_{L}^\text{emg}$ at the last MoME layer $L$ are outputted.

\textbf{Mix Pathway.}
The mix pathway serves to capture both intra-modal and cross-modal information by concatenating tokens from EEG and EMG and feeding them to the MoME layers. Initially, the mix tokens $\boldsymbol{T}_{0}^\text{mix}$ pass through MSA in the early MoME layers before being split and mapped back to their original latent space to maintain modality coherence. Using EEG/EMG modality experts ensures stable modality interaction, as earlier layers' low-level features are susceptible to noise. By projecting representations to their own space, the potential noise entering another space is constrained to a certain degree. However, at later layers, token representations are mapped to a multi-modal space with mixed experts to enable deeper modality interaction. The MoME module's output is the token series from the last layer, denoted as $\boldsymbol{T}_{L}^\text{mix}$.

\subsection{Epoch Mixture-of-Modality-Experts Transformer}
Following epoch settings in previous works~\cite{phan2018joint, pradeepkumar2022towards}, we proposed an Epoch sDREAMER model for sleep state classification with a one-to-one paradigm. The Epoch sDREAMER, depicted in Fig.~\ref{fig:mome}, consists of an epoch-level MoME module with three pathways, each assigned a modality-specific classification head. The model takes the initial representations for patches of a given epoch signal as input, and the epoch-level MoME module aggregates representative information among patches to mine epoch context.

During training, all three pathways are enabled. The EEG and EMG initial representations propagate along their respective pathways to produce the final representations, and their corresponding ${\left[\text{CLS}\right]}$ tokens $\boldsymbol{T}_\text{cls}^\text{eeg}$ and $\boldsymbol{T}_\text{cls}^\text{emg}$ are fed to their respective classifiers for prediction. However, the model still lacked cross-modal information, so we leveraged the mix pathway to learn rich cross-modal information while maintaining the information within each modality. The final $\boldsymbol{T}_\text{cls}^\text{mix}$ token is utilized, which is a combination of the EEG and EMG representations. During inference, the model can flexibly select the pathway according to the specific token type and generate predictions using only the relevant information within the input token.

\subsection{Sequence Mixture-of-Modality-Experts Transformer}
To tackle sleep staging in a sequence setting, we proposed a sequence sDREAMER model that captures contextual information hierarchically. 
 Unlike the epoch setting, where each epoch is considered as a separate sequence, Sequence sDREAMER views each epoch in relation to its neighboring epochs. The architecture of the proposed Sequence sDREAMER model is illustrated in Fig.~\ref{fig:Seq_MoME}, which employs an epoch-level MoME module at a lower hierarchy to capture epoch-level contexts. Next, a sequence-level MoME transformer is used to capture sequence-level contexts at a higher hierarchy.

\textbf{Epoch Encoder.}
Although the epoch-level MoME module in Epoch sDREAMER and Sequence sDREAMER transformer shared the same network architecture, their information flow differs for both training and inference stages. The mix pathway is not enabled in the epoch-level MoME module because multi-modality information is better modeled during the sequence stage at a higher level. This design decision ensures robustness to noise during training and speeds up inference. Thus, the epoch-level MoME module functions as a context extractor for EEG and EMG epoch signals.
To compute the initial representations for an EEG/EMG sequence consisting of $K$ epochs, we concatenate the initial tokens from each epoch to form the whole sequence's epoch-level initial representations, denoted as $\boldsymbol{H}_{0}^{m}$. Each epoch's initial representation is then passed through the EEG/EMG pathway. Finally, the resulting $\left[\text{CLS}\right]$ tokens that have passed through the MoME module for all $K$ epochs are concatenated to form the sequence-level initial representations of the given modality.
\begin{align}
    \boldsymbol{H}_{0}^{m} &= \left[\boldsymbol{T}_{0}^{m(1)}; \boldsymbol{T}_{0}^{m(2)}; \cdots ; \boldsymbol{T}_{0}^{m(K)}\right] \\
    \boldsymbol{T}_\text{cls}^{m(i)} &= \text{Epoch-MoME}\left(\boldsymbol{T}_{0}^{m(i)}\right),  i = 1 \ldots K \\
    \boldsymbol{Z}_{0}^{m} &= \text{Concat}\left(\boldsymbol{T}_\text{cls}^{m(1)}; \boldsymbol{T}_\text{cls}^{m(2)}; \cdots ;\boldsymbol{T}_\text{cls}^{m(K)} \right) 
\end{align}
After passing through both pathways, two sequence-level tokens are generated: $\boldsymbol{Z}_{0}^\text{eeg}$ and $\boldsymbol{Z}_{0}^\text{emg}$. 

\textbf{Sequence Encoder.}
Considering a series of EEG/EMG tokens outputted from the epoch-level MoME module as the context features of each epoch, we now transform them into initial representations to sequence MoME module. To denote the temporal relationship between epochs,  two sequence-level positional embeddings $\boldsymbol{E}_\text{pos}^\text{eeg} \in \mathbb{R}^{K \times D}$ and $\boldsymbol{E}_\text{pos}^\text{emg} \in \mathbb{R}^{K \times D}$ are added elementwise to EEG and EMG tokens, respectively. \\
Next, the sequence-level MoME blocks begin to learn dependencies between epochs within the same sequence. 
Like the Epoch sDREAMER, the EEG and EMG pathways process the input tokens separately and map their learned representations to modality-specific latent space. The output tokens from each pathway are then projected with pathway-specific classification heads to the label space. Regarding the mix pathway, the mixed EEG-EMG tokens are first projected separately before being mapped to a multi-modal latent space. To better aggregate the multi-modal information learned in the mix pathway, the output EEG-EMG tokens from the same epoch are concatenated along the feature dimension. These concatenated tokens are then mapped to the label space with another pathway-specific head. Finally, three sets of output predictions are generated: $\left\{\boldsymbol{z}_{i}^\text{eeg}\right\}_{i=1}^{K}$, $\left\{\boldsymbol{z}_{i}^\text{emg}\right\}_{i=1}^{K}$ and $\left\{\boldsymbol{z}_{i}^\text{mix}\right\}_{i=1}^{K}$. \\

\subsection{Self-Distillation}
Provided a multi-pathways structure, we seek an effective approach that enables such pathways to co-improve with each other. 
To this end, a self-distillation framework for better mono-modal and multi-modal representation learning is presented. 
To specify, the output from the mix pathway is leveraged to distill the outputs from EEG and EMG pathways. Given the outputs of each pathway, the self-distilled loss formulates as the KL-divergence between the predictions from EEG/EMG and pseudo-targets provided by the mixed pathway. 
Let $\boldsymbol{p}_\tau$ denotes the softmax function with temperature factor $\tau$ defined as 
\begin{align}
    \boldsymbol{p}_{\tau}(\boldsymbol{z}_i^{m(k)}) = \frac{\text{exp}(\boldsymbol{z}_i^{m(k)} / \tau)}{\sum_{j}^{C}\text{exp}(\boldsymbol{z}_i^{m(j)} / \tau)},
\end{align}
where $z_i^{m(k)}$ denotes the $k$-th dimension of the logits vector for $i$-th prediction given by $m$ modality pathway. The self-distillation loss for EEG and EMG formulates as shown in Equation \ref{eq:self-distill-eeg} and \ref{eq:self-distill-emg}, respectively.

\begin{equation}
\label{eq:self-distill-eeg}
\begin{aligned}
     \mathcal{L}_\text{sd-eeg} = \text{KL}\left(\boldsymbol{p}_{\tau_\text{eeg}}\left(\boldsymbol{z}^\text{eeg}\right) \parallel \boldsymbol{p}_{\tau_\text{eeg}}\left(\boldsymbol{z}^\text{mix}\right)\right) 
\end{aligned}
\end{equation}
\begin{equation}
\label{eq:self-distill-emg}
\begin{aligned}
     \mathcal{L}_\text{sd-emg} = \text{KL}\left(\boldsymbol{p}_{\tau_\text{emg}}\left(\boldsymbol{z}^\text{emg}\right) \parallel \boldsymbol{p}_{\tau_\text{emg}}\left(\boldsymbol{z}^\text{mix}\right)\right) 
\end{aligned}
\end{equation}
The logits vectors of predictions from three pathways are denoted as $\boldsymbol{z}^\text{eeg}$, $\boldsymbol{z}^\text{emg}$, and $\boldsymbol{z}^\text{mix}$. $\tau_\text{eeg}$ and $\tau_{emg}$ are temperature factors for the distillation of EEG and EMG. Along with the self-distillation loss, a cross-entropy loss is also incorporated into the optimization process.
\begin{equation}
\label{eq:CE}
\begin{aligned}
    \mathcal{L}_\text{ce} &= {-\sum_{i=1}^{L} \boldsymbol{y}_i\log(\boldsymbol{p}_{\tau=1}(\boldsymbol{z}_{i}^\text{mix}))}
\end{aligned}
\end{equation}
The $\boldsymbol{y}_i$ and $\boldsymbol{z}_i^\text{mix}$ indicate the $i$-th ground truth and prediction from the mix pathway. $L$ is the number of epoch signals.
The total loss for MoME models, shown in Equation \ref{eq:MoME}, is a linear combination of self-distillation and cross-entropy loss.
\begin{equation}
\label{eq:MoME}
\begin{aligned}
    \mathcal{L}_\text{mome} = \left(1-\alpha\right)\mathcal{L}_\text{ce} + \frac{\alpha}{2}\left(\mathcal{L}_\text{sd-emg} + \mathcal{L}_\text{sd-emg}\right)
\end{aligned}
\end{equation}
Here, the $\alpha$ is a scaling factor used to balance the weightage of the loss terms.

\section{Experiments}

\begin{table*}[t]
\centering
\caption{Performance comparison between different model architectures on the 
mouse sleep staging dataset.}
{
\begin{tabular}{ccccccc}
\toprule
\textbf{Method}                          & \textbf{Hierarchy} & \textbf{Shared Self-Attention} & \textbf{CA Level} & \textbf{\# of Epochs} & \textbf{Acc(\%)}            & \textbf{F1-Score(\%)}       \\ \midrule
Decision Tree                                  & Epoch      & -            & -          & 1             & 79.25          & 67.65          \\
Random Forest                                  & Epoch      & -            & -          & 1             & 85.47          & 74.55          \\
AdaBoost                                       & Epoch      & -            & -          & 1             & 85.85          & 77.78          \\
XGBoost\cite{chen2016xgboost}                                        & Epoch      & -            & -          & 1             & 85.29          & 75.79          \\
MLP                                            & Epoch      & -            & -          & 1             & 81.46          & 67.05          \\
Bi-LSTM                                        & Epoch      & -            & -          & 1             & 85.57          & 76.09          \\
Channel-Independent Transformer\cite{nie2022time}& Epoch      & \checkmark   & -          & 1             & 86.70          & 78.40          \\
Dual-Encoder Transformer                           & Epoch      & -            & -          & 1             & 87.17          & 78.68          \\
Cross-Attention Transformer                    & Epoch      & -            & Epoch level      & 1             & 87.07          & 79.13          \\
Cross-Modal Transformer\cite{pradeepkumar2022towards} & Epoch      & -            & -          & 1             & 87.01          & -              \\ \midrule
Epoch sDREAMER                                      & Epoch      & \checkmark   & -          & 1             & \textbf{88.25} & \textbf{81.30} \\ \midrule
MLP + Transformer                              & Sequence   & -            & -          & 16            & 88.88          & 83.33          \\
Bi-LSTM                                        & Sequence   & -            & -          & 16            & 90.07          & 85.21          \\
Dual-Encoder Transformer                           & Sequence   & -            & -          & 16            & 90.65          & 86.37          \\
Cross-Attention Transformer                    & Sequence   & -            & Epoch level      & 16            & 90.37          & 86.31          \\
SeqCross-Attention Transformer                 & Sequence   & -            & Sequence level      & 16            & 90.93          & 86.48          \\
Cross-Modal Transformer\cite{pradeepkumar2022towards} & Sequence   & -            & Sequence level      & 16            & 87.84          & -          \\ \midrule
Sequence sDREAMER                                        & Sequence   & \checkmark   & -       & 16            & \textbf{91.72} & \textbf{87.64} \\ \bottomrule
\end{tabular}}
\label{tab:method comparison}
\end{table*}

\subsection{Experiment Setup}
\label{sec:dataset}
\begin{figure}[htbp]
    \centering
    \includegraphics[width=0.48\textwidth]{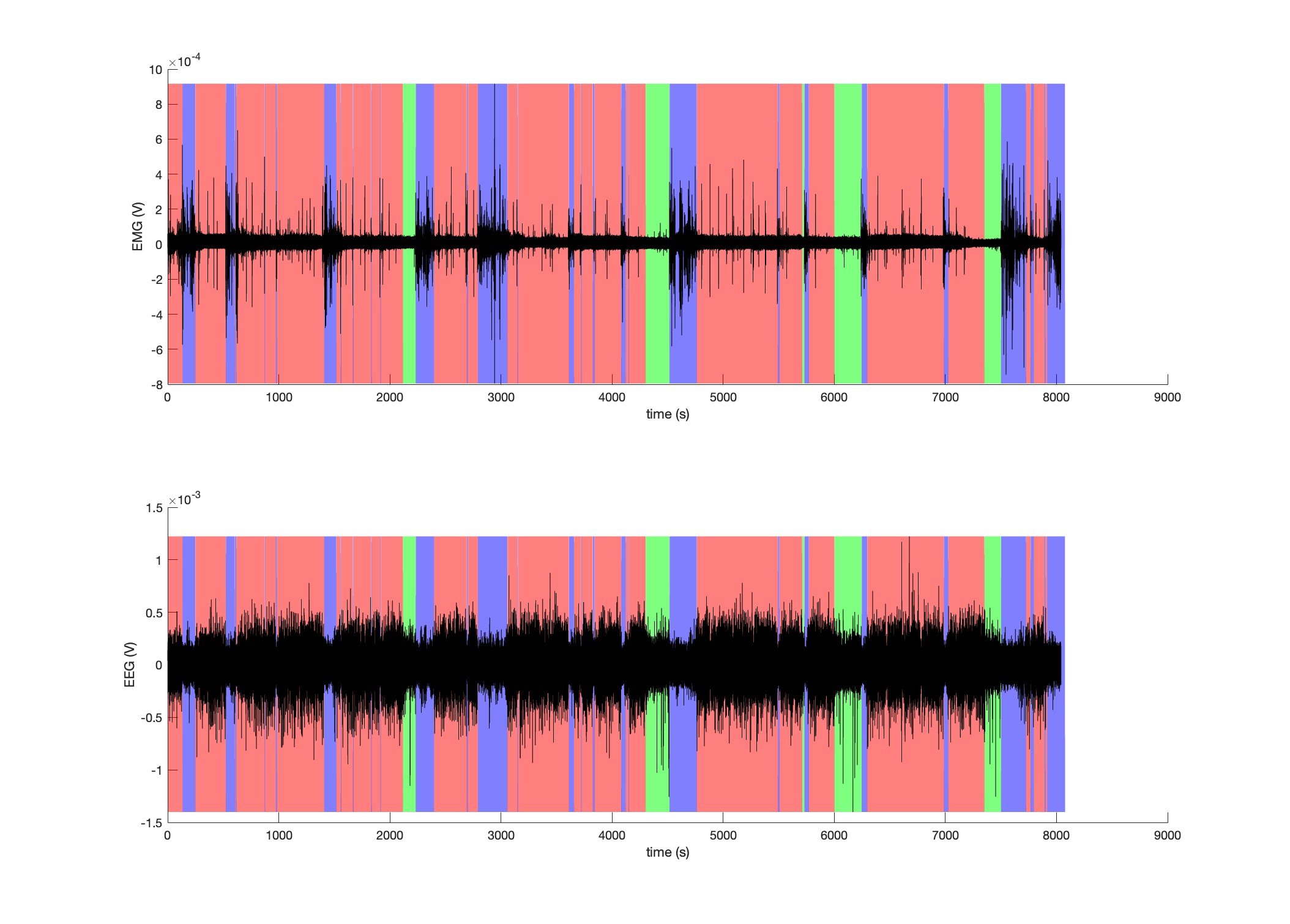}
    \caption{An example of the collected raw data. The two graphs represent the EMG and EEG data, respectively. The background colors of blue, green, and red are the labels of sleep stages.}
    \label{fig:data}
\end{figure}

\textbf{Dataset. }
We conduct an evaluation of our proposed epoch sDREAMER and sequence sDREAMER models using the 
mouse sleep staging dataset, which so far comprises EEG-EMG paired signals collected from mice during sleep and will include additional modalities in the future. Experts have labeled the sleep stages of the mice every second based on these electrophysiological signals, assigning each second of recorded data one of the following labels: Wake, SWS (slow-wave sleep), or REM (rapid eye movement). An illustrative example of the data is shown in Fig.~\ref{fig:data}.

\begin{figure}[htbp]
    \centering
	\begin{minipage}[t]{0.45\linewidth}
    \includegraphics[width=1.0\textwidth]{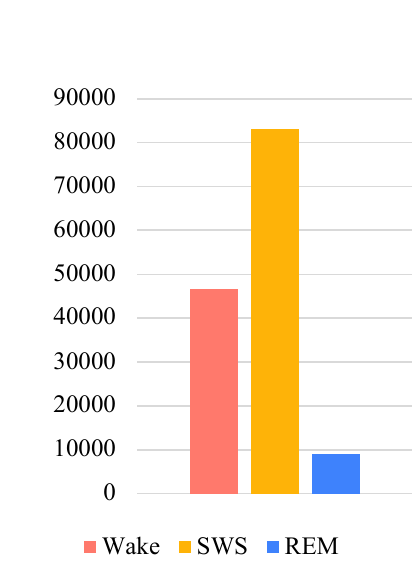}
    \subcaption{Bar plot of the training set}
    \label{fig:bar_train}
    \end{minipage}
    \hfill
    \begin{minipage}[t]{0.45\linewidth}
    \includegraphics[width=1.0\textwidth]{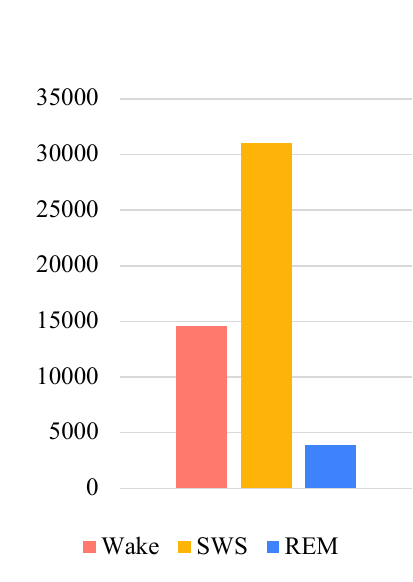}
    \subcaption{Bar plot of the test set}
    \label{fig:bar_test}
    \end{minipage}
    \caption{Bar plot of the class distribution, including (a) training set and (b) test set.}\label{fig:bar_total}
\end{figure}

The dataset consists of EEG and EMG records of 16 mouse subjects, with each record spanning approximately 4 hours. Given that the EEG and EMG signals are sampled at 512 Hz, there can be about 5-10 million EEG/EMG data points for each mouse record. To align with the expert labeling span, we set the epoch window to 1 second, resulting in a total of 10,000 epoch data samples. To evaluate the performance of our model, we adopted subject-wise split criteria, where the records of 12 mice subjects were used for training, while the records of 4 mice subjects were used for testing. The distribution of stages in the training set and test set can be seen in Fig.~\ref{fig:bar_train} and Fig.~\ref{fig:bar_test}, respectively.

\subsection{Implementation Details}
In this section, we explain the epoch and sequence settings of our sDREAMER model. The epoch MoME model involves a 4-layer MoME module, with the first three layers having only EEG and EMG experts, and the last layer incorporating a mixed expert. The feed-forward network dimension for each expert is 512. The Sequence sDREAMER model comprises a 2-layer MoME module for epoch modeling and a 3-layer MoME module for sequence modeling. The entire epoch-level MoME module and the first two layers of the sequence MoME module have only EEG and EMG experts, and the final layer of the sequence MoME module has an additional mix expert.
All experiments are performed on an NVIDIA RTX 3090 GPU with AdamW\cite{loshchilov2017decoupled} as the optimizer. The learning rate and weight decay are set to 1e-3 and 1e-4, respectively. The non-hierarchical model utilizes a batch size of 256, while the hierarchical model uses a batch size of 16. The EEG and EMG self-distillation temperature factors are set to $\tau_\text{eeg}=1.0$ and $\tau_\text{emg}=3.0$, respectively, and the distillation weight $\alpha$ is 0.33.

\subsection{Baselines} 
We compare our method with both machine learning-based and deep learning-based methods. Due to the limited work for mice sleep staging, we implemented multiple baseline model architectures with different settings, from simple to complex. We also include a state-of-the-art cross-modal transformer method~\cite{pradeepkumar2022towards} as one of the baselines.

\textbf{Machine-learning-based Methods.}
Decision tree, AdaBoost, XGBoost, and Random forest were chosen as the baselines for the machine learning-based methods. However, due to the high dimensionality of the EEG-EMG signal, capturing the underlying patterns within the raw signal is challenging for these methods. Therefore, we engineered several features of EEG-EMG traces. In particular, statistical features(\eg\ mean, max, min, std, and skew) for the raw signal and its first-order derivative are leveraged. Additionally, we applied a discrete Fourier transform to both the raw signal and its derivative and incorporated the statistics in the frequency domain as well. 

\textbf{Deep-learning-based Methods.}
For deep-learning methods, we developed several baseline models and also compared others' methods. To start with, MLP is a straightforward model with multiple feedforward layers. Bi-LSTM is a sequential modeling model that captures temporal contexts from the forward and backward directions.
 The channel-independent transformer~\cite{nie2022time} is an architecture that shares the attention weight for multi-channel input signals.
A dual-encoder transformer represents a typical late-fusion model that uses two independent encoders for each input signal and merges the information late in the architecture. The cross-attention transformer is a variant of the dual-encoder architecture that add the cross-attention layer to capture cross-modal information. Lastly, the state-of-the-art cross-modal transformer~\cite{pradeepkumar2022towards}, another variant of the dual-encoder model, uses a self-attention layer to integrate information from two modalities.

\begin{table*}[t]
    \centering
    \caption{Mono-modal inference performance comparison of the vanilla transformer and Sequence sDREAMER on the mouse sleep dataset. The results with the best performance are highlighted in bold. "EEG/EMG-Acc" represents the prediction accuracy when taking EEG/EMG as input. Similarly, "EEG/EMG-F1" represents the F1-score in the same context.}
    \label{tab:mono-modality}
    \begin{tabular}{cccccc}
    \toprule
    \textbf{Method}              & \textbf{Train Signal}        & \textbf{EEG-Acc(\%)} & \textbf{EEG-F1(\%)} & \textbf{EMG-Acc(\%)} & \textbf{EMG-F1(\%)} \\ \midrule
    \multicolumn{1}{c|}{Dual-Encoder Transformer} & \multicolumn{1}{c|}{EEG} & 87.16 & 72.32 & -              & -     \\
    \multicolumn{1}{c|}{Dual-Encoder Transformer} & \multicolumn{1}{c|}{EMG} & -     & -     & \textbf{83.78} & 57.57 \\ 
    \multicolumn{1}{c|}{Sequence sDREAMER} & \multicolumn{1}{c|}{EEG-EMG} & \textbf{88.12}       & \textbf{77.83}      & \textbf{83.78}       & \textbf{62.82}      \\ \bottomrule
    \end{tabular}
\end{table*}

\subsection{Quantitative Results}
\textbf{Multi-modal Comparison.}
In order to demonstrate the efficacy of our model, we conducted a comparative analysis against the aforementioned baseline methods on the multi-modal input setting. Table~\ref{tab:method comparison} presents the experiment results for EEG-EMG paired input, where we report the accuracy and F1-score for each model. Results show that our proposed epoch-level sDREAMER outperforms the machine-learning-based methods by a significant margin. Furthermore, our sDREAMER model outperforms all the deep-learning-based methods, including the state-of-the-art method cross-modal transformer, on both epoch level and sequence level.

\textbf{Mono-modal Comparison.}
To demonstrate our model's generalizability on single-modal inputs, we evaluated our model's ability to produce accurate sleep stage predictions using a single EEG or EMG signal input in the inference stage. It is noteworthy that none of the baseline methods trained with multi-modal input can make inference with single-modal input. Therefore, we compare our model's performance with the baseline dual-encoder transformer model trained solely on EEG or EMG. Results are presented in Table~\ref{tab:mono-modality}. The results indicate that our proposed model was able to learn a superior mono-modal representation by utilizing the multi-modality as a form of supervision during training.

\begin{table}[t] \scriptsize
\caption{Ablation study on modality-experts self-distillation. The accuracy is reported. }
\label{tab:ablation-distillation}
\begin{tabular}{cccccc}
\toprule
\textbf{Method}                                 & \textbf{EEG-SD} & \textbf{EMG-SD}        & \textbf{Acc} & \textbf{EEG-Acc} & \textbf{EMG-Acc} \\ \midrule      
\multicolumn{1}{c|}{Sequence sDREAMER} &  \checkmark & \multicolumn{1}{c|}{\checkmark}           & \textbf{91.72} & \textbf{88.12} & \textbf{83.78} \\ \midrule
\multicolumn{1}{c|}{\multirow{3}{*}{Ablations}} &  \multicolumn{1}{c}{\checkmark}    & \multicolumn{1}{c|}{-} & 91.45            & 87.57                & 22.67                \\
\multicolumn{1}{c|}{}        & - & \multicolumn{1}{c|}{\checkmark}           & 90.02          & 60.18 & 83.49 \\
\multicolumn{1}{c|}{}        & - & \multicolumn{1}{c|}{\textbf{-}} & 77.13          & 62.71 & 7.85  \\ \bottomrule
\end{tabular}
\end{table}

\subsection{Ablation Studies}
In this section, we present the ablation studies for our model. Specifically, as the effectiveness of the MoME transformer has been validated, we hereby conduct an ablation study regarding positional and modality embedding to investigate the contribution of different attribute embedding. As shown in Table~\ref{tab:ablation-encoding}, the joint attribute encoding improves the performance by an increase of approximately $1.0\%$ in accuracy. Among the two encodings, positional encoding is of greater importance than modality encoding. One possible explanation for this is that there are only two kinds of electrophysiological signals incorporated within this sleep staging task, thus weakening the impacts of modality encoding. In addition, we also perform an ablation study on the contribution of self-distillation. Table~\ref{tab:ablation-distillation} indicates that the self-distillation not only improved the mono-modality performance significantly but also promoted the multi-modal performance by a large margin.
\begin{table}[t]
\caption{Ablation study on positional encoding and modality encoding. The accuracy is reported. }
\label{tab:ablation-encoding}
\begin{tabular}{cccc}
\toprule
\textbf{Method}                                 & \textbf{Pos Encoding} & \textbf{Mod Encoding}           & \textbf{Acc(\%)} \\ \midrule
\multicolumn{1}{c|}{Sequence sDREAMER}                    & \checkmark                      & \multicolumn{1}{c|}{\checkmark}           & \textbf{91.72}   \\ \midrule
\multicolumn{1}{c|}{\multirow{3}{*}{Ablations}} & -                     & \multicolumn{1}{c|}{\checkmark}           & 91.45            \\
\multicolumn{1}{c|}{}                           & \checkmark                      & \multicolumn{1}{c|}{-}          & 91.65            \\
\multicolumn{1}{c|}{}                           & -                     & \multicolumn{1}{c|}{\textbf{-}} & 90.99            \\ \bottomrule
\end{tabular}
\end{table}

\begin{figure*}[htbp]
    \centering
    \includegraphics[width=0.9\linewidth]{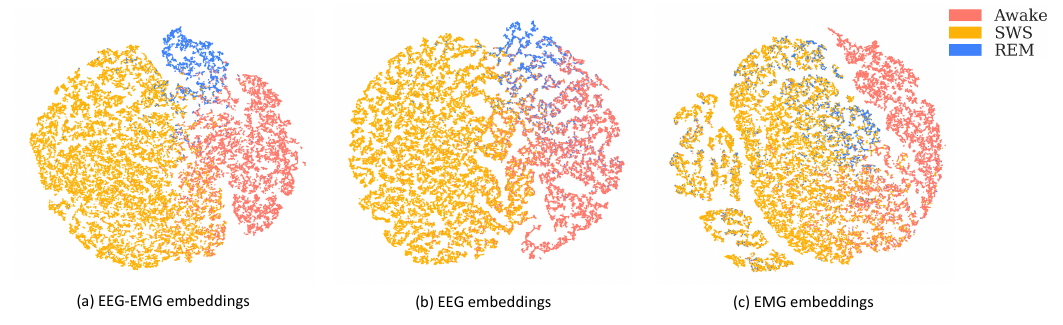}
    \caption{Visualization of learned embeddings using t-SNE.}
    \label{fig:tsne}
\end{figure*}

\begin{figure*}[htbp]
    \centering
    \includegraphics[width=0.75\linewidth]{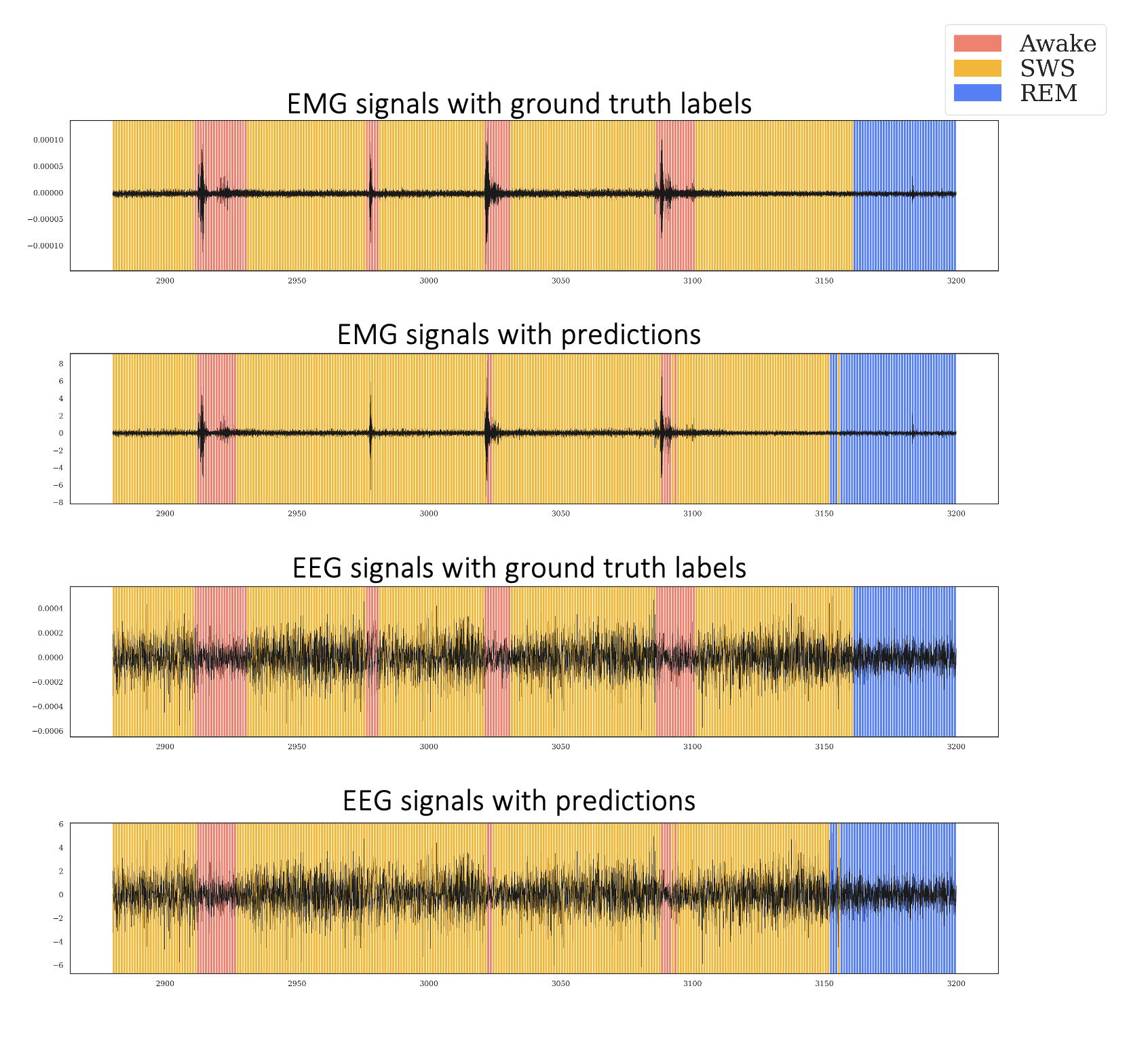}
    \caption{Visualization of the Predictions given by the sDREAMER.}
    \label{fig:Sequence sDREAMER pred}
\end{figure*}

\subsection{Visualizations}
\textbf{t-SNE Analysis.}
To explore the embeddings learned by our model, we utilized t-SNE~\cite{van2008visualizing} to visualize the high-dimensional embedding in a two-dimensional space. The resulting visualization, shown in Fig.~\ref{fig:tsne}, demonstrates the disentangled and aligned latent space learned by Sequence sDREAMER. 
In particular, the representations spaces for all three pathways are visualized, and they indicate a similar spatial pattern.
This pattern alignment is non-trivial because we did not provide any supervision directly that would promote such alignment in the latent space\textemdash suggesting shared attention mechanism and self-distillation have implicitly benefited the multi-modal alignment.
The t-SNE plot also reveals spatial continuity among the Wake, REM, and SWS sleep stages in the embedding space, consistent with real-world scenarios. Our model implicitly learns to maintain the relative spatial relationships in the ground truth space without injecting any prior knowledge of actual spatial relativity. These findings highlight the effectiveness and interpretability of our proposed method in sleep stage scoring.

\textbf{Visualizations.}
In Fig.~\ref{fig:Sequence sDREAMER pred}, we give some predictions results of our sequence sDREAMER model. From the visualization results, we see that our model predicts most of the sleep stages correct. 
\section{Conclusions and Discussions}
Automatic sleep staging is a crucial first step for sleep-related research. In this paper, we propose sDREAMER, a sleep staging model that can handle both single-channel and multi-channel inputs. Our sDREAMER emphasizes the cross-modality interaction
and per-channel performance, and therefore generates high-quality staging results on the sleep staging task. Extensive experiments demonstrate the effectiveness of our proposed sDREAMER.

There are some limitations in our work. The performance of the model is currently compared to one expert, who is well-trained but not perfect. Our model may show greater potential if the dataset is scored by more experts. In addition, if the dataset is scored by two newly trained students, this experiment will show whether the model can outperform newly trained students and therefore replace human scoring.

\section*{Acknowledgments}
Research reported in this publication was supported by the National Institutes of Health under Award Number U19NS128613. The content is solely the responsibility of the authors and does not necessarily represent the official views of the National Institutes of Health.
{\small
\bibliographystyle{ieee_fullname}
\bibliography{egbib}

\begin{thebibliography}{10}\itemsep=-1pt

\bibitem{akada2021deep}
Keishi Akada, Takuya Yagi, Yuji Miura, Carsten~T Beuckmann, Noriyuki Koyama, and Ken Aoshima.
\newblock A deep learning algorithm for sleep stage scoring in mice based on a multimodal network with fine-tuning technique.
\newblock {\em Neuroscience Research}, 173:99--105, 2021.

\bibitem{allen2020towards}
Zeyuan Allen-Zhu and Yuanzhi Li.
\newblock Towards understanding ensemble, knowledge distillation and self-distillation in deep learning.
\newblock {\em arXiv preprint arXiv:2012.09816}, 2020.

\bibitem{andonian2022robust}
Alex Andonian, Shixing Chen, and Raffay Hamid.
\newblock Robust cross-modal representation learning with progressive self-distillation.
\newblock In {\em Proceedings of the IEEE/CVF Conference on Computer Vision and Pattern Recognition}, pages 16430--16441, 2022.

\bibitem{andreotti2018multichannel}
Fernando Andreotti, Huy Phan, Navin Cooray, Christine Lo, Michele~TM Hu, and Maarten De~Vos.
\newblock Multichannel sleep stage classification and transfer learning using convolutional neural networks.
\newblock In {\em 2018 40th annual international conference of the IEEE engineering in medicine and biology society (EMBC)}, pages 171--174. IEEE, 2018.

\bibitem{bao2022vlmo}
Hangbo Bao, Wenhui Wang, Li Dong, Qiang Liu, Owais~Khan Mohammed, Kriti Aggarwal, Subhojit Som, Songhao Piao, and Furu Wei.
\newblock Vlmo: Unified vision-language pre-training with mixture-of-modality-experts.
\newblock {\em Advances in Neural Information Processing Systems}, 35:32897--32912, 2022.

\bibitem{biswal2017sleepnet}
Siddharth Biswal, Joshua Kulas, Haoqi Sun, Balaji Goparaju, M~Brandon Westover, Matt~T Bianchi, and Jimeng Sun.
\newblock Sleepnet: automated sleep staging system via deep learning.
\newblock {\em arXiv preprint arXiv:1707.08262}, 2017.

\bibitem{chen2016xgboost}
Tianqi Chen and Carlos Guestrin.
\newblock Xgboost: A scalable tree boosting system.
\newblock In {\em Proceedings of the 22nd acm sigkdd international conference on knowledge discovery and data mining}, pages 785--794, 2016.

\bibitem{dong2022maskclip}
Xiaoyi Dong, Yinglin Zheng, Jianmin Bao, Ting Zhang, Dongdong Chen, Hao Yang, Ming Zeng, Weiming Zhang, Lu Yuan, Dong Chen, et~al.
\newblock Maskclip: Masked self-distillation advances contrastive language-image pretraining.
\newblock {\em arXiv preprint arXiv:2208.12262}, 2022.

\bibitem{dosovitskiy2020image}
Alexey Dosovitskiy, Lucas Beyer, Alexander Kolesnikov, Dirk Weissenborn, Xiaohua Zhai, Thomas Unterthiner, Mostafa Dehghani, Matthias Minderer, Georg Heigold, Sylvain Gelly, et~al.
\newblock An image is worth 16x16 words: Transformers for image recognition at scale.
\newblock {\em arXiv preprint arXiv:2010.11929}, 2020.

\bibitem{eldele2021attention}
Emadeldeen Eldele, Zhenghua Chen, Chengyu Liu, Min Wu, Chee-Keong Kwoh, Xiaoli Li, and Cuntai Guan.
\newblock An attention-based deep learning approach for sleep stage classification with single-channel eeg.
\newblock {\em IEEE Transactions on Neural Systems and Rehabilitation Engineering}, 29:809--818, 2021.

\bibitem{hou2019learning}
Yuenan Hou, Zheng Ma, Chunxiao Liu, and Chen~Change Loy.
\newblock Learning lightweight lane detection cnns by self attention distillation.
\newblock In {\em Proceedings of the IEEE/CVF international conference on computer vision}, pages 1013--1021, 2019.

\bibitem{huang2022improved}
Jing Huang, Lifeng Ren, Xiaokang Zhou, and Ke Yan.
\newblock An improved neural network based on senet for sleep stage classification.
\newblock {\em IEEE Journal of Biomedical and Health Informatics}, 26(10):4948--4956, 2022.

\bibitem{jia2021scaling}
Chao Jia, Yinfei Yang, Ye Xia, Yi-Ting Chen, Zarana Parekh, Hieu Pham, Quoc Le, Yun-Hsuan Sung, Zhen Li, and Tom Duerig.
\newblock Scaling up visual and vision-language representation learning with noisy text supervision.
\newblock In {\em International Conference on Machine Learning}, pages 4904--4916. PMLR, 2021.

\bibitem{jia2022multi}
Ziyu Jia, Xiyang Cai, and Zehui Jiao.
\newblock Multi-modal physiological signals based squeeze-and-excitation network with domain adversarial learning for sleep staging.
\newblock {\em IEEE Sensors Journal}, 22(4):3464--3471, 2022.

\bibitem{jia2021salientsleepnet}
Ziyu Jia, Youfang Lin, Jing Wang, Xuehui Wang, Peiyi Xie, and Yingbin Zhang.
\newblock Salientsleepnet: Multimodal salient wave detection network for sleep staging.
\newblock {\em arXiv preprint arXiv:2105.13864}, 2021.

\bibitem{kidger2020neural}
Patrick Kidger, James Morrill, James Foster, and Terry Lyons.
\newblock Neural controlled differential equations for irregular time series.
\newblock {\em Advances in Neural Information Processing Systems}, 33:6696--6707, 2020.

\bibitem{kim2018automatic}
Hyungjik Kim and Sunwoong Choi.
\newblock Automatic sleep stage classification using eeg and emg signal.
\newblock In {\em 2018 Tenth International Conference on Ubiquitous and Future Networks (ICUFN)}, pages 207--212. IEEE, 2018.

\bibitem{kjaerby2022memory}
Celia Kjaerby, Mie Andersen, Natalie Hauglund, Verena Untiet, Camilla Dall, Bj{\"o}rn Sigurdsson, Fengfei Ding, Jiesi Feng, Yulong Li, Pia Weikop, et~al.
\newblock Memory-enhancing properties of sleep depend on the oscillatory amplitude of norepinephrine.
\newblock {\em Nature neuroscience}, 25(8):1059--1070, 2022.

\bibitem{li2021align}
Junnan Li, Ramprasaath Selvaraju, Akhilesh Gotmare, Shafiq Joty, Caiming Xiong, and Steven Chu~Hong Hoi.
\newblock Align before fuse: Vision and language representation learning with momentum distillation.
\newblock {\em Advances in neural information processing systems}, 34:9694--9705, 2021.

\bibitem{loshchilov2017decoupled}
Ilya Loshchilov and Frank Hutter.
\newblock Decoupled weight decay regularization.
\newblock {\em arXiv preprint arXiv:1711.05101}, 2017.

\bibitem{mikkelsen2018personalizing}
Kaare Mikkelsen and Maarten De~Vos.
\newblock Personalizing deep learning models for automatic sleep staging.
\newblock {\em arXiv preprint arXiv:1801.02645}, 2018.

\bibitem{miladinovic2019spindle}
{\DJ}or{\dj}e Miladinovi{\'c}, Christine Muheim, Stefan Bauer, Andrea Spinnler, Daniela Noain, Mojtaba Bandarabadi, Benjamin Gallusser, Gabriel Krummenacher, Christian Baumann, Antoine Adamantidis, et~al.
\newblock Spindle: End-to-end learning from eeg/emg to extrapolate animal sleep scoring across experimental settings, labs and species.
\newblock {\em PLoS computational biology}, 15(4):e1006968, 2019.

\bibitem{mobahi2020self}
Hossein Mobahi, Mehrdad Farajtabar, and Peter Bartlett.
\newblock Self-distillation amplifies regularization in hilbert space.
\newblock {\em Advances in Neural Information Processing Systems}, 33:3351--3361, 2020.

\bibitem{nie2022time}
Yuqi Nie, Nam~H Nguyen, Phanwadee Sinthong, and Jayant Kalagnanam.
\newblock A time series is worth 64 words: Long-term forecasting with transformers.
\newblock {\em arXiv preprint arXiv:2211.14730}, 2022.

\bibitem{PATHAK2021102038}
Shreyasi Pathak, Changqing Lu, Sunil~Belur Nagaraj, Michel {van Putten}, and Christin Seifert.
\newblock Stqs: Interpretable multi-modal spatial-temporal-sequential model for automatic sleep scoring.
\newblock {\em Artificial Intelligence in Medicine}, 114:102038, 2021.

\bibitem{phan2018automatic}
Huy Phan, Fernando Andreotti, Navin Cooray, Oliver~Y Ch{\'e}n, and Maarten De~Vos.
\newblock Automatic sleep stage classification using single-channel eeg: Learning sequential features with attention-based recurrent neural networks.
\newblock In {\em 2018 40th annual international conference of the IEEE engineering in medicine and biology society (EMBC)}, pages 1452--1455. IEEE, 2018.

\bibitem{phan2018dnn}
Huy Phan, Fernando Andreotti, Navin Cooray, Oliver~Y Ch{\'e}n, and Maarten De~Vos.
\newblock Dnn filter bank improves 1-max pooling cnn for single-channel eeg automatic sleep stage classification.
\newblock In {\em 2018 40th annual international conference of the IEEE engineering in medicine and biology society (EMBC)}, pages 453--456. IEEE, 2018.

\bibitem{phan2018joint}
Huy Phan, Fernando Andreotti, Navin Cooray, Oliver~Y Ch{\'e}n, and Maarten De~Vos.
\newblock Joint classification and prediction cnn framework for automatic sleep stage classification.
\newblock {\em IEEE Transactions on Biomedical Engineering}, 66(5):1285--1296, 2018.

\bibitem{phan2019seqsleepnet}
Huy Phan, Fernando Andreotti, Navin Cooray, Oliver~Y Ch{\'e}n, and Maarten De~Vos.
\newblock Seqsleepnet: end-to-end hierarchical recurrent neural network for sequence-to-sequence automatic sleep staging.
\newblock {\em IEEE Transactions on Neural Systems and Rehabilitation Engineering}, 27(3):400--410, 2019.

\bibitem{phan2021xsleepnet}
Huy Phan, Oliver~Y Ch{\'e}n, Minh~C Tran, Philipp Koch, Alfred Mertins, and Maarten De~Vos.
\newblock Xsleepnet: Multi-view sequential model for automatic sleep staging.
\newblock {\em IEEE Transactions on Pattern Analysis and Machine Intelligence}, 2021.

\bibitem{phan2022sleeptransformer}
Huy Phan, Kaare Mikkelsen, Oliver~Y Ch{\'e}n, Philipp Koch, Alfred Mertins, and Maarten De~Vos.
\newblock Sleeptransformer: Automatic sleep staging with interpretability and uncertainty quantification.
\newblock {\em IEEE Transactions on Biomedical Engineering}, 69(8):2456--2467, 2022.

\bibitem{pradeepkumar2022towards}
Jathurshan Pradeepkumar, Mithunjha Anandakumar, Vinith Kugathasan, Dhinesh Suntharalingham, Simon~L Kappel, Anjula~C De~Silva, and Chamira~US Edussooriya.
\newblock Towards interpretable sleep stage classification using cross-modal transformers.
\newblock {\em arXiv preprint arXiv:2208.06991}, 2022.

\bibitem{radford2021learning}
Alec Radford, Jong~Wook Kim, Chris Hallacy, Aditya Ramesh, Gabriel Goh, Sandhini Agarwal, Girish Sastry, Amanda Askell, Pamela Mishkin, Jack Clark, et~al.
\newblock Learning transferable visual models from natural language supervision.
\newblock In {\em International conference on machine learning}, pages 8748--8763. PMLR, 2021.

\bibitem{schwabedal2018automated}
Justus~TC Schwabedal, Daniel Sippel, Moritz~D Brandt, and Stephan Bialonski.
\newblock Automated classification of sleep stages and eeg artifacts in mice with deep learning.
\newblock {\em arXiv preprint arXiv:1809.08443}, 2018.

\bibitem{shen2022self}
Yiqing Shen, Liwu Xu, Yuzhe Yang, Yaqian Li, and Yandong Guo.
\newblock Self-distillation from the last mini-batch for consistency regularization.
\newblock In {\em Proceedings of the IEEE/CVF Conference on Computer Vision and Pattern Recognition}, pages 11943--11952, 2022.

\bibitem{sors2018convolutional}
Arnaud Sors, St{\'e}phane Bonnet, S{\'e}bastien Mirek, Laurent Vercueil, and Jean-Fran{\c{c}}ois Payen.
\newblock A convolutional neural network for sleep stage scoring from raw single-channel eeg.
\newblock {\em Biomedical Signal Processing and Control}, 42:107--114, 2018.

\bibitem{stephansen2018neural}
Jens~B Stephansen, Alexander~N Olesen, Mads Olsen, Aditya Ambati, Eileen~B Leary, Hyatt~E Moore, Oscar Carrillo, Ling Lin, Fang Han, Han Yan, et~al.
\newblock Neural network analysis of sleep stages enables efficient diagnosis of narcolepsy.
\newblock {\em Nature communications}, 9(1):5229, 2018.

\bibitem{supratak2017deepsleepnet}
Akara Supratak, Hao Dong, Chao Wu, and Yike Guo.
\newblock Deepsleepnet: A model for automatic sleep stage scoring based on raw single-channel eeg.
\newblock {\em IEEE Transactions on Neural Systems and Rehabilitation Engineering}, 25(11):1998--2008, 2017.

\bibitem{tan2019lxmert}
Hao Tan and Mohit Bansal.
\newblock Lxmert: Learning cross-modality encoder representations from transformers.
\newblock {\em arXiv preprint arXiv:1908.07490}, 2019.

\bibitem{van2008visualizing}
Laurens Van~der Maaten and Geoffrey Hinton.
\newblock Visualizing data using t-sne.
\newblock {\em Journal of machine learning research}, 9(11), 2008.

\bibitem{vaswani2017attention}
Ashish Vaswani, Noam Shazeer, Niki Parmar, Jakob Uszkoreit, Llion Jones, Aidan~N Gomez, {\L}ukasz Kaiser, and Illia Polosukhin.
\newblock Attention is all you need.
\newblock {\em Advances in neural information processing systems}, 30, 2017.

\bibitem{vilamala2017deep}
Albert Vilamala, Kristoffer~H Madsen, and Lars~K Hansen.
\newblock Deep convolutional neural networks for interpretable analysis of eeg sleep stage scoring.
\newblock In {\em 2017 IEEE 27th international workshop on machine learning for signal processing (MLSP)}, pages 1--6. IEEE, 2017.

\bibitem{wang2022image}
Wenhui Wang, Hangbo Bao, Li Dong, Johan Bjorck, Zhiliang Peng, Qiang Liu, Kriti Aggarwal, Owais~Khan Mohammed, Saksham Singhal, Subhojit Som, et~al.
\newblock Image as a foreign language: Beit pretraining for all vision and vision-language tasks.
\newblock {\em arXiv preprint arXiv:2208.10442}, 2022.

\bibitem{wang2021distilled}
Zekun Wang, Wenhui Wang, Haichao Zhu, Ming Liu, Bing Qin, and Furu Wei.
\newblock Distilled dual-encoder model for vision-language understanding.
\newblock {\em arXiv preprint arXiv:2112.08723}, 2021.

\bibitem{yang2019snapshot}
Chenglin Yang, Lingxi Xie, Chi Su, and Alan~L Yuille.
\newblock Snapshot distillation: Teacher-student optimization in one generation.
\newblock In {\em Proceedings of the IEEE/CVF Conference on Computer Vision and Pattern Recognition}, pages 2859--2868, 2019.

\bibitem{zhang2021cross}
Dingwen Zhang, Guohai Huang, Qiang Zhang, Jungong Han, Junwei Han, and Yizhou Yu.
\newblock Cross-modality deep feature learning for brain tumor segmentation.
\newblock {\em Pattern Recognition}, 110:107562, 2021.

\bibitem{zhang2019your}
Linfeng Zhang, Jiebo Song, Anni Gao, Jingwei Chen, Chenglong Bao, and Kaisheng Ma.
\newblock Be your own teacher: Improve the performance of convolutional neural networks via self distillation.
\newblock In {\em Proceedings of the IEEE/CVF International Conference on Computer Vision}, pages 3713--3722, 2019.

\bibitem{zhang2021vinvl}
Pengchuan Zhang, Xiujun Li, Xiaowei Hu, Jianwei Yang, Lei Zhang, Lijuan Wang, Yejin Choi, and Jianfeng Gao.
\newblock Vinvl: Revisiting visual representations in vision-language models.
\newblock In {\em Proceedings of the IEEE/CVF Conference on Computer Vision and Pattern Recognition}, pages 5579--5588, 2021.

\bibitem{zhang2021perturbed}
Yachao Zhang, Yanyun Qu, Yuan Xie, Zonghao Li, Shanshan Zheng, and Cuihua Li.
\newblock Perturbed self-distillation: Weakly supervised large-scale point cloud semantic segmentation.
\newblock In {\em Proceedings of the IEEE/CVF International Conference on Computer Vision}, pages 15520--15528, 2021.

\bibitem{zhang2020self}
Zhilu Zhang and Mert Sabuncu.
\newblock Self-distillation as instance-specific label smoothing.
\newblock {\em Advances in Neural Information Processing Systems}, 33:2184--2195, 2020.

\bibitem{zhao2021knowledge}
Haoran Zhao, Xin Sun, Junyu Dong, Zihe Dong, and Qiong Li.
\newblock Knowledge distillation via instance-level sequence learning.
\newblock {\em Knowledge-Based Systems}, 233:107519, 2021.

\end{thebibliography}
}

\end{document}